\documentclass[letterpaper, 10 pt, conference]{ieeeconf}  % Comment this line out if you need a4paper

\IEEEoverridecommandlockouts                              % This command is only needed if 
\usepackage{cite}
\usepackage{graphics}
\usepackage{epsfig}
\usepackage{booktabs}
\usepackage{amsmath,amssymb,amsfonts}

\title{\LARGE \bf
A Generalized Control Revision Method for Autonomous Driving Safety
}

\author{Zehang Zhu$^{1,\#}$, Yuning Wang$^{1,2,\#}$, Tianqi Ke$^{1}$, Zeyu Han$^{1}$, Shaobing Xu$^{1}$, \\
Qing Xu$^{1,*}$, John M. Dolan$^{2}$, and Jianqiang Wang$^{1}$% <-this % stops a space
\thanks{This work was supported by National Key R\&D Program of China (2023YFB2504400) and the National Natural Science Foundation of China under Grant 52221005.}% <-this % stops a space
\thanks{\#These authors contribute to this work equally.}% <-this % stops a space
\thanks{*Corresponding author: Qing Xu.}% <-this % stops a space
\thanks{$^{1}$The authors are with the School of Vehicle and Mobility, Tsinghua University, Beijing, China.
        {\tt\small \{zzh22, wangyn20, ktq23, hanzy21\}@mails.tsinghua.edu.cn, \{shaobxu, qingxu, wjqlws\}@tsinghua.edu.cn}}%
\thanks{$^{2}$The authors are with the Robotics Institute, Carnegie Mellon University, Pittsburgh, PA 15213 USA.
        {\tt\small jdolan@andrew.cmu.edu}}%
}

\begin{document}

\maketitle
\thispagestyle{empty}
\pagestyle{empty}

%%%%%%%%%%%%%%%%%%%%%%%%%%%%%%%%%%%%%%%%%%%%%%%%%%%%%%%%%%%%%%%%%%%%%%%%%%%%%%%%
\begin{abstract}

Safety is one of the most crucial challenges of autonomous driving vehicles, and one solution to guarantee safety is to employ an additional control revision module after the planning backbone. Control Barrier Function (CBF) has been widely used because of its strong mathematical foundation on safety. However, the incompatibility with heterogeneous perception data and incomplete consideration of traffic scene elements make existing systems hard to be applied in dynamic and complex real-world scenarios. In this study, we introduce a generalized control revision method for autonomous driving safety, which adopts both vectorized perception and occupancy grid map as inputs and comprehensively models multiple types of traffic scene constraints based on a new proposed barrier function. Traffic elements are integrated into one unified framework, decoupled from specific scenario settings or rules. Experiments on CARLA, SUMO, and OnSite simulator prove that the proposed algorithm could realize safe control revision under complicated scenes, adapting to various planning backbones, road topologies, and risk types. Physical platform validation also verifies the real-world application feasibility.

\end{abstract}

%%%%%%%%%%%%%%%%%%%%%%%%%%%%%%%%%%%%%%%%%%%%%%%%%%%%%%%%%%%%%%%%%%%%%%%%%%%%%%%%
\section{INTRODUCTION}

Safety is one of the main challenges for autonomous driving (AD) technologies to be implemented on real-road applications~\cite{yang2018grand}, especially for vehicle planning and control. Although extensive studies have employed various situation awareness methods in autonomous driving planning modules to address the safety issue~\cite{wang2023decision}, the uncertainty and complexity of real-world road conditions still could result in scenarios that deviate from model assumptions, leading to collisions in practical applications~\cite{grahn2020expert}. Both Hierarchical rule-based planning systems and end-to-end methods can still not guarantee safety well~\cite{schwarting2018planning, aradi2020survey, wang2024survey}.

\begin{figure}
	\centering
	\includegraphics[width=8.5cm]{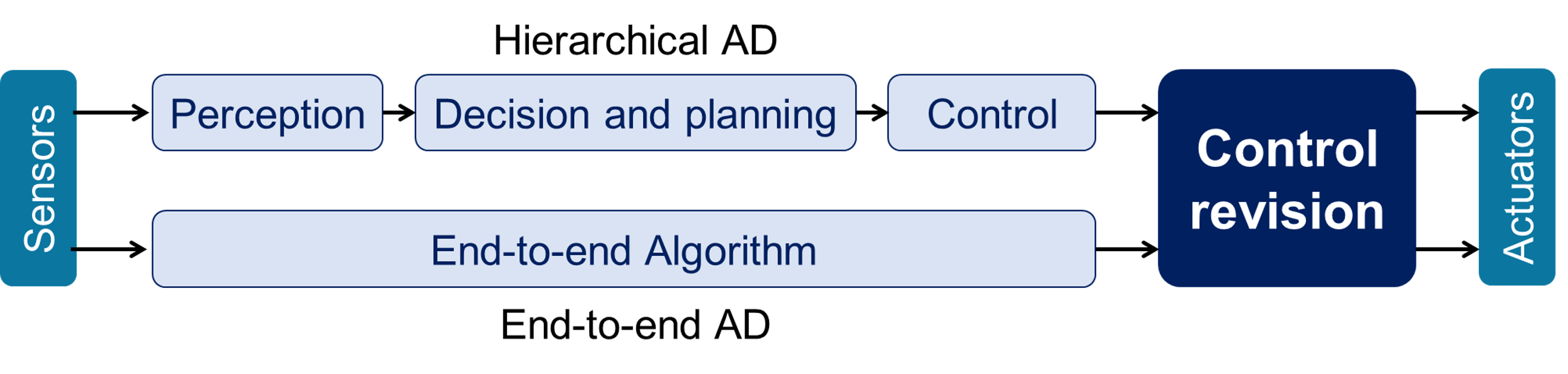}
	\caption{A diagram of the control revision module.}  
	\label{Fig:Safety definition}
\end{figure}

One solution to ensure the driving safety is to employ an addition control revision module after planning, as shown in Fig. \ref{Fig:Safety definition}. The concept of control revision is widely used in various kinds of AD systems especially in recent end-to-end AD planning frameworks~\cite{wang2023drivemlm, hu2023planning}, defined as a rule-based module revising the original control output to avoid accident. Control revision should be totally white-box since it is the last module towards interaction with the real world. 

Control Barrier Function (CBF) has attracted wide interest for safe control because of its mathematical guarantee in control stability and absolute safety~\cite{ames2016control}. Some studies have applied CBF in AD control. Lyu et al. utilized CBF in a bi-level optimization framework to realize safety-assured adaptive merging control under ramp scenarios~\cite{lyu2021probabilistic}. He et al. applied CBF on racing vehicles attached after a Model Predictive Control planning algorithm to avoid collision in enclosing racing tracks~\cite{he2022autonomous}. Alan implemented CBF on a class-8 truck and tested the static obstacle avoidance~\cite{alan2023control}.

Although CBF has been proven to be effective in control revision for safety, two challenges remain. The first problem is the incompatibility with heterogeneous perception input formats~\cite{jian2023dynamic}. Most CBF-based control methods took vectorized bounding boxes as inputs to represent the features of obstacles, which are naturally appropriate for CBF mathematical expressions. However, apart from vectorized bounding box, occupancy grid map is also a commonly used perception output format. It remains a challenge to cope with heterogeneous data so that the original scenario can be translated in a unified format and understandable for CBF. The second issue is the inadaptability to the dynamic and complicated traffic scenario elements~\cite{liu2023safety}. As reviewed above, most current studies only considered a single kind of obstacles under specific scenarios such as ramps, highways, etc., lacking modeling other risk sources including road restrictions, complex dynamic agent interactions, etc~\cite{wang2023emergency}. To ensure the generalization ability of CBF control revision module, a unified method is needed to transfer various perception results and model the scenario safety risk comprehensively.

In this study, we propose a generalized control revision method facing complicated and dynamic driving scenarios to ensure safety. We aim to achieve a unified CBF control framework which adopts various kinds of perception information and considers different types of traffic risk sources. The main contributions are as follows:
\begin{itemize}
    \item A perception data conversion layer that provides uniform scenario inputs for CBF, adopting heterogeneous perception results;
    \item A unified traffic element constraint layer considering both dynamic agent risks and static road topology;
    \item Demonstration of the effectiveness on multiple simulators and real-world platforms.
\end{itemize}

The following paper is organized as follows. Section~\ref{Section 2} gives the preliminaries of CBF-related definitions. Section~\ref{Section 3} introduces the details of perception conversion, constraint representation, and CBF integration. Experiments and validations are demonstrated in Section~\ref{Section 4}. Section~\ref{Section 5} provides the conclusions.

\section{Preliminaries and background}
\label{Section 2}

In this section, we briefly introduce CBF and related work for further applications. 
Consider a close-loop nonlinear affine control system:
\begin{equation}
    \dot{\boldsymbol{x}} = f(\boldsymbol{x}) + g(\boldsymbol{x})\boldsymbol{u}
    \label{system}
\end{equation}
where $ \boldsymbol{x} \in \mathbb{X} \subseteq \mathbb{R}^n $ is state vector, 
$ \boldsymbol{u} \in \mathbb{U} \subseteq \mathbb{R}^m $ is control input. 
Define the set $C$ as the superlevel set of a continuously differentiable function $h: \mathbb{X} \rightarrow \mathbb{R}$:
\begin{equation}
    C = \{\boldsymbol{x} \in \mathbb{X}: h(\boldsymbol{x}) \geqslant 0 \}
    \label{safeset}
\end{equation}
We denote $C$ as the safe set and $h$ as the barrier function throughout this paper. 
Referring to CBF theory, to keep state $\boldsymbol{x}$ staying in the safe set, the control input $\boldsymbol{u} \in \mathbb{U}$ should satisfy the CBF constraint:
\begin{equation}
    L_f h + L_g h \boldsymbol{u} + \alpha(h) \geqslant 0
    \label{cbf_ori}
\end{equation}
where $ L_f h$ and $ L_g h$ denote the Lie derivatives of $ h $ along $ f $ and $ g $ in system \eqref{system} respectively, 
and $\alpha$ is a selected extended class $\mathcal{K}_{\infty}$ function.

\emph{Dynamic Control Barrier Function (D-CBF) \cite{jian2023dynamic}}: 
Consider a movable obstacle, whose state vector $ \boldsymbol{x}_{ob} \in \mathbb{X}_{ob} \subseteq \mathbb{R}^p $ and the barrier function $h: \mathbb{X} \times \mathbb{X}_{ob} \rightarrow \mathbb{R}$.
To maintain state $\boldsymbol{x}$ in the safe set, the control input $\boldsymbol{u} \in \mathbb{U}$ should satisfy the D-CBF constraint:
\begin{equation}
    L_f h + L_g h \boldsymbol{u} + \frac{\partial h}{\partial \boldsymbol{x}_{ob}} \dot{\boldsymbol{x}}_{ob} + \alpha(h) \geqslant 0
    \label{cbf_dy}
\end{equation}

\emph{High Order Control Barrier Function (HOCBF) \cite{xiao2019control}}: 
For the barrier function with the relative degree $r$ with respect to system \eqref{system}, 
where $r := \min\{ q \in \mathbb{N}:  \frac{\partial }{\partial \boldsymbol{u} } \frac{\partial^q h}{\partial t^q } \neq 0, \exists \boldsymbol{x} \in \mathbb{X} \}$, it is found that $L_g h = 0$ if $r > 1$, 
so that a CBF-based optimization problem cannot be formulated since the control vector $\boldsymbol{u}$ doesn't show up in \eqref{cbf_ori}. 
By differentiating the barrier function along system \eqref{cbf_ori} $r$ times, the constraint containing $\boldsymbol{u}$ can be inferred as:
\begin{equation}
    L_f^r h + L_g L_f^{r-1} h \boldsymbol{u} + \sum_{i=0}^{r-1} L_f^i(\alpha_{r-i}(h_{r-i-1})) \geqslant 0
    \label{hocbf}
\end{equation}
where $h_0 := h$, and $h_i:= \dot{h_{i-1}+\alpha_i(h_{i-1})}$ for $i = \{1,2,...,r\}$; 
$\alpha_i(\cdot)$ is a $(r-i)^{th}$ order differentiable extended class $\mathcal{K}_{\infty}$ function for $i = \{1,2,...,r\}$; 
$L_f^i h$ denotes the Lie derivatives of $ h $ along $ f $ for $ i $ times.

\emph{Feasibility Constraint \cite{xiao2022sufficient}}:
For physical system, control limits always exist and may conflict with the CBF constraints presented above. To avoid confliction, define the feasibility constraint as:
\begin{equation}
    h_F = L_f h + L_g h \boldsymbol{u^*} + \alpha(h) \geqslant 0
    \label{fs}
\end{equation}
where $\boldsymbol{u^*}$ presents control vector that maximizes LHS in \eqref{cbf_ori} under the physical limits. 
Stare $\boldsymbol{x}$ can stay in the safe set only if $h_F \geqslant 0$.
Consider \eqref{fs} as a new barrier function, an actuator-complied CBF constraint can be formulated.

\section{Methods}
\label{Section 3}

\begin{figure}
	\centering
	\includegraphics[width=8.5cm]{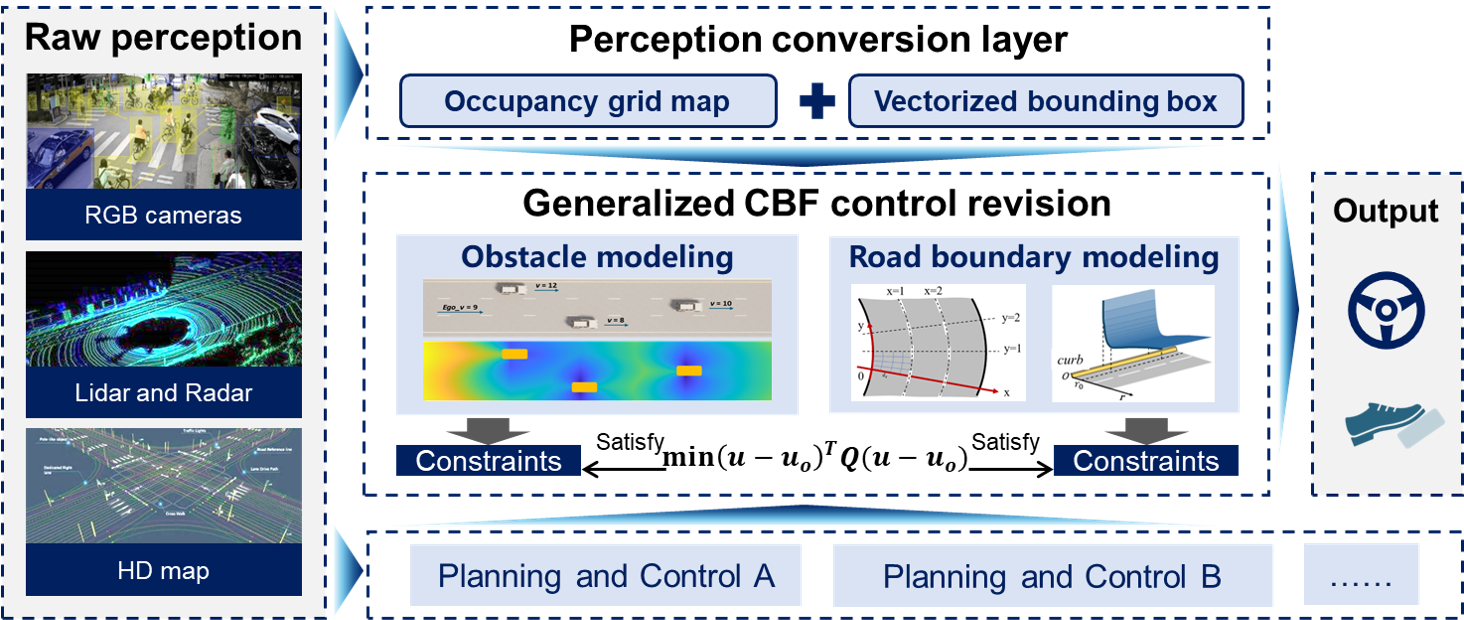}
	\caption{The framework of the proposed control revision method.}  
	\label{Fig:framework}
\end{figure}

The framework of the proposed control revision method is demonstrated in Fig.~\ref{Fig:framework}. The perception conversion layer transfers heterogeneous raw data into one unified manner. Then with the vanilla planning results from upstream module and the converted perception, the CBF-based system revise the control to ensure safety by considering obstacle and road boundary constraints.

\subsection{Perception Data Conversion} 
\label{section percetpion conversion}
\begin{figure}
	\centering
	\includegraphics[width=8.5cm]{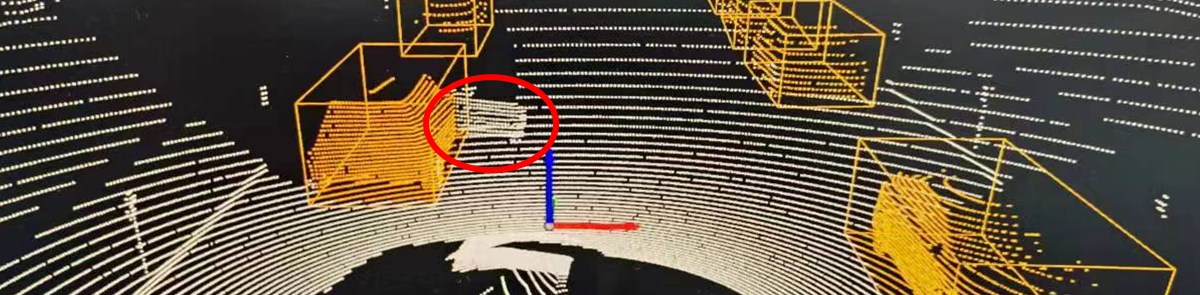}
	\caption{Failure of bounding box to represent irregular shaped obstacles.}  
	\label{Fig:vector problem}
\end{figure}

To enhance the practicability of safe control revision, it's important to adopt heterogeneous perception results formats. Current AD perception output can be mainly divided into two categories: vectorized bounding box and occupancy grid map (OGM), depending on the sensor used for data collection ~\cite{velasco2020autonomous}. Camera-based perception generations vectorized bounding boxes which represent the obstacles by vectors, as expressed in \eqref{bounding box def} where $(x_i,y_i)$ is the coordinate of the $i^{th}$ obstacle center, $l_i$ is the length, $w_i$ is the width, and $\theta_i$ is the heading angle. One problem of vision-based perception outputs is that the defined bounding box is a standard cube, which makes it struggling to represent irregular-shaped obstacles. Fig.~\ref{Fig:vector problem} gives a failure example of the bounding box to represent irregular shaped obstacles. In this scene, a parked car opened a door while the perception method still only recognized the main vehicle body.

\begin{equation}
    \label{bounding box def}
    obs = [[x_1,y_1,l_1,w_1, \theta_1],...]
\end{equation}

Another category is OGM, mainly generated by Lidar or Radar-based perception methods~\cite{li2020deep}. Some recent studies also developed networks to predict OGM by RGB cameras~\cite{wei2023surroundocc}. The scenario is divided into several small rectangular grids, each of which can be in one of two states: occupied or unoccupied. Different from vectorized bounding box, OGM is not constrained by specific shapes so that it can represent obstacles with irregular agents well. However, because it does not emphasize the integrity of individual obstacles, this type of format is comparatively more difficult to be understood by the planning and control module.

% \begin{figure}
% 	\centering
% 	\includegraphics[width=8.5cm]{Figures/OGM-v2.png}
% 	\caption{A diagram of occupancy grid map.}  
% 	\label{Fig:OGM def}
% \end{figure}

For a control revision that aims to ensure absolute safety, perception completeness is crucial. Even though CBF can guarantee mathematical safety proof, if the obstacle inputs are missing, accidents will still happen. In addition, to make the revision module more generalized, being compatible with heterogeneous perception formats is also important. In this study, we design a perception data conversion layer which uses the OGM as supplementary inputs for vectorized bounding box to make the traffic scene representation more complete and understandable for CBF.

The core idea is to represent the undetected grids by new supplementary bounding boxes. Fig.~\ref{Fig: supple perception} is a diagram of the supplementary perception conversion procedures. First, using the detected vectorized bounding boxes to delete the grids that have been covered so that the left ones are obstacles need complements. Normally, the left grids are the irregular extending parts of existing obstacles, so their density and numbers would not be large in one scene. Then hierarchical clustering, which does not require pre-inputing the cluster number~\cite{zhang2023review}, is used to separate the remaining occupied grids into neighboring groups. For each cluster, we can derive a minimum convex hull that covers all grids by Graham scanning algorithm~\cite{an2021modified}. The last step is to solve the minimum rectangle, which can be expressed by a five-tuple bounding box defined in~\eqref{bounding box def}, for each convex hull by enumerating all minimum rectangles that share an edge with the hull and comparing their area. As shown in Fig.~\ref{Fig: supple perception}, the supplementary bounding boxes are attached to the original vectorized perception results and served as the obstacle inputs of CBF system. Note that this process is still feasible with only OGM as inputs, equivalent to the situation where the vectorized results are none.

\begin{figure}
	\centering
	\includegraphics[width=8.5cm]{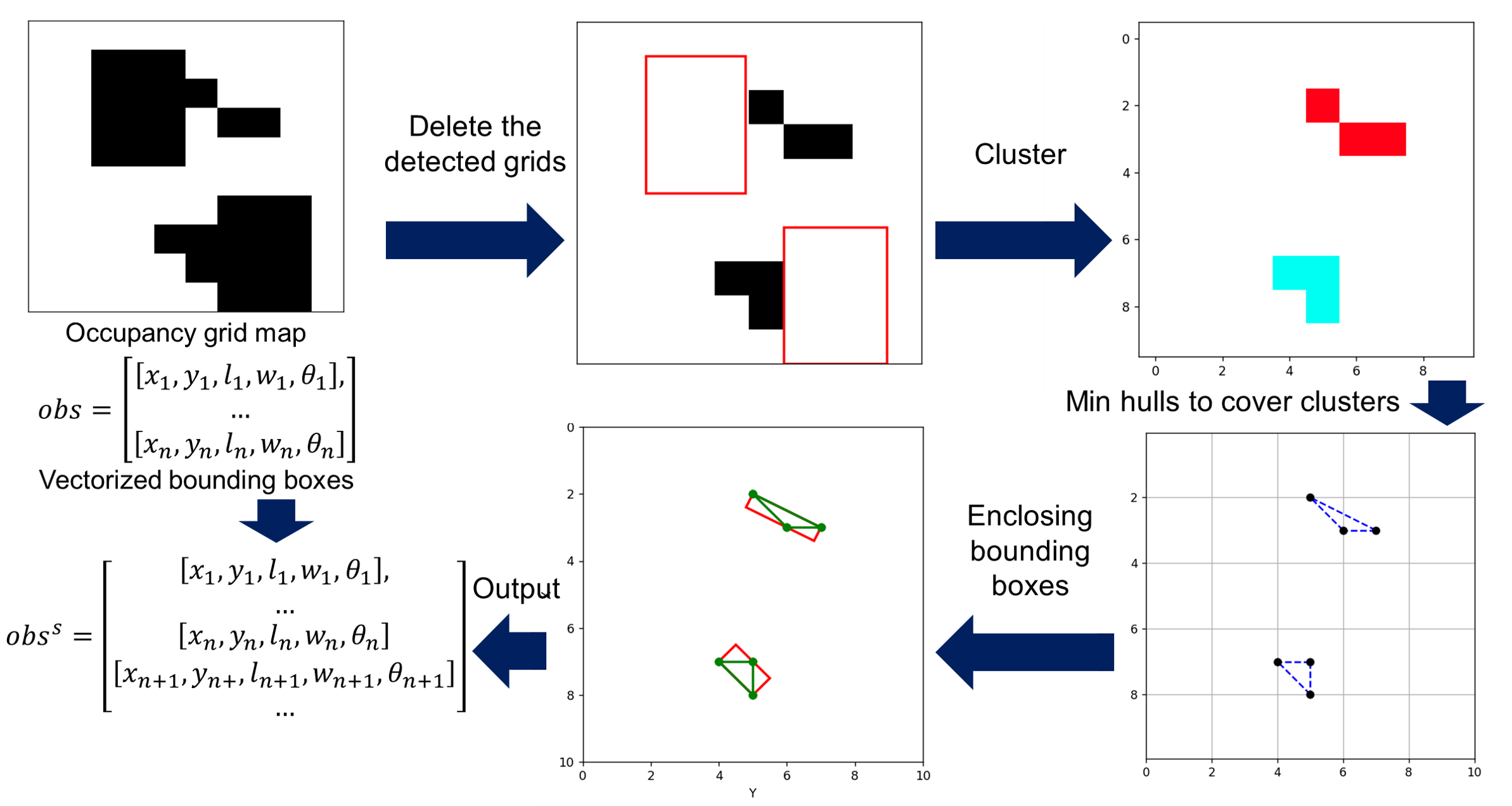}
	\caption{A diagram of supplementary perception conversion procedures.}  
	\label{Fig: supple perception}
\end{figure}

Through the proposed perception conversion layer, the completeness and generalization ability of control revision is enhanced.

\subsection{Traffic Element Constraint Representation}
\label{section cbf constraint}

Unlike most robot systems and open scenarios, the complexity of the dynamic scenes generated from vehicle characteristics, traffic participants, and regulations set up challenges to CBF's applications to AD. In this part, we introduce the designs of the barrier function and constraints in our safe revision method.

\emph{1) Vehicle Dynamic Characteristics}

We use the vehicle kinematic model as ego vehicle control system. 
To simplify later usage, slip angle is assumed as a small angle since the lateral acceleration of the vehicle is small. %(in this paper, longitudinal direction is the orientation the same with vehicle heading, and lateral direction is the orientation perpendicular to longitudinal direction).
The system is described as follows:
\begin{equation}
	\begin{split}
		\dot{\boldsymbol{x}}
		& = \left[
			\begin{array}{cc}
				v \cos(\varphi) \\
				v \sin(\varphi) \\
				0 \\
				0 \\
			\end{array}
		\right] +
		\left[
			\begin{array}{cc}
				0 & 0 \\
				0 & 0 \\
				1 & 0 \\
				0 & v / L \\
			\end{array}
		\right]
		\left[
			\begin{array}{cc}
				a \\
				\tan(\delta_f) \\
			\end{array}
		\right] \\
		& = f(\boldsymbol{x}) + g(\boldsymbol{x}) \boldsymbol{u} .
	\end{split}
	\label{veh system}
\end{equation}
where $\boldsymbol{x} = [x_g, y_g, v, \varphi]^T$ is the state vector, and $ \boldsymbol{u} = [a, \tan(\delta_f)]^T $ is the control input;
$ x_g $ and $ y_g $ denote the coordinates of ego vehicle's mass center;
$ v $ and $ a $ describe the longitudinal speed and acceleration of the ego vehicle;
$ \varphi $ represents the orientation of the ego vehicle, also as ego vehicle heading;
$ \delta_f $ describes the front steering angle;
$ L $ and $ l_r $ describes the wheelbase and rear wheelbase respectively, which are constant parameters for a certain vehicle.
Note that system \eqref{veh system} is an affine system, since control input $ \boldsymbol{u} = [a, \tan(\delta_f)]^T $ instead of $[a, \delta_f]^T $.

An important dynamic characteristic of vehicle system need to be considered is the asymmetry between longitudinal and lateral behaviors. 
From system \eqref{veh system}, the vehicle can't move laterally only since the change of vehicle heading relating with the longitudinal speed, and it may cause extra risk if the vehicle has a large lateral offset in traffic environments.
So, different from using Euclidean distance or considering only direction to build the barrier function in many related work \cite{jian2023dynamic}\cite{liu2023safety}, we propose a new kind of barrier function for the vehicle system based on longitudinal and lateral distances:
\begin{equation}
    h = \sqrt[]{\frac{d^2_{lon}}{l^2_{lon}} + \frac{d^2_{lat}}{l^2_{lat}}} - c_{safe}\\
    \label{bf}
\end{equation}
where $ d_{lon} $ and $ d_{lat} $ are functions of ego vehicle position, ego vehicle heading and the obstacle position, describing the longitudinal and lateral distance of the obstacle in ego vehicle coordinate;
$ l_{lon} $ and $ l_{lat} $ presents the longitudinal and lateral safety coefficients, normally $l_{lon} > l_{lat}$, and $ c_{safe} $ is a safety constant.
In our work, we define $ l_{lon} $ and $ l_{lat} $ are functions of headings and shapes of the ego vehicle and the obstacle.

The barrier function \eqref{bf} is also suitable for movable obstacles. Moreover, in \eqref{bf} ego vehicle coordinate won't rotate when the ego vehicle steers, avoiding steering frequently.
In this way, the barrier function \eqref{bf} has the relative degree 2 with respect to system \eqref{veh system}. %, since only positions in state vector are contained in the barrier function \eqref{bf} as variables.

\emph{2) Constraints Refinement for Traffic Participates}

Following the barrier function \eqref{bf}, based on D-CBF\cite{jian2023dynamic} and HOCBF\cite{xiao2019control} we replace origin CBF constraints with D-CBF constraints and take its derivative. Selecting linear functions as extended class $\mathcal{K}_{\infty}$ functions, the constraints can be inferred as:
\begin{equation}
    \begin{split}
        &B(\boldsymbol{x}, \boldsymbol{u}|h) = L_f^2 h + L_g L_f h \boldsymbol{u} 
        + 2 \frac{\partial L_f h}{\partial \boldsymbol{x}_{ob}}\dot{\boldsymbol{x}}_{ob} \\
        &~~~~~~~+ \dot{\boldsymbol{x}}_{ob}^T \frac{\partial^2 h}{\partial \boldsymbol{x}_{ob}^2} \dot{\boldsymbol{x}}_{ob} 
        + (\alpha_1 + \alpha_2) \dot{h} + \alpha_1 \alpha_2 h \geqslant 0          
    \end{split}
    \label{d cbf}
\end{equation}

Vehicle actuator constrains the applied control input on vehicle system, providing maximum and minimum limits of control input of system \eqref{veh system}. 
Besides, road fiction sets dynamic constraints of vehicle system, and the envelope of the curves of longitudinal force and lateral force is an ellipse, called attachment ellipse. 
According to these constraints, we calculate control input $\boldsymbol{u^*}$ that maximize the barrier function $B(\boldsymbol{x}, \boldsymbol{u})$ \eqref{d cbf}, and have the feasibility constraint \cite{xiao2022sufficient} for integrating CBF:
\begin{equation}
    \begin{split}
        h_F = \underset{\boldsymbol{u^*} \in \mathbb{U}}{\max} B(\boldsymbol{x}, \boldsymbol{u^*}|h) \geqslant 0          
    \end{split}
    \label{fs cbf}
\end{equation}
Then, the CBF constraint could be built based on the feasibility constraint \eqref{fs cbf}.

\textbf{Remark: }
\emph{Further discussion of vehicle control limits would be held, since attachment ellipse is an approximation. In this paper, we use maximum or minimum acceleration with zero steer as $\boldsymbol{u^*}$ to avoid infeasible constraint.}

\emph{3) Road Boundary Constraint}

Although most traffic regulations have been considered in decision part of autonomous driving, lacking consideration of traffic regulation may cause unreasonable even unsafe control input after safe revision in traffic environment. 
In this part, we introduce the modeling of road boundary constraint.

In above parts, our discussions are mainly held in rectangular coordinate system, which is intuitive to analyze the behavior of the ego vehicle and obstacles, but it is hard to handle road boundary. 
To get a convex constraint, we consider Frenet coordinate, with small slip angle assumption, in which the vehicle system is:
\begin{equation}
    \left[
        \begin{array}{cc}
            \dot{s} \\
            \dot{d} \\
            \dot{v} \\
            \dot{\mu} \\
        \end{array}
    \right]
    = \left[
        \begin{array}{cc}
            \frac{v \cos(\mu)}{1 - d\kappa} \\
            v \sin(\mu) \\
            0 \\
            -\kappa \frac{v \cos(\mu)}{1 - d\kappa} \\
        \end{array}
    \right]  +
    \left[
        \begin{array}{cc}
            0 & 0 \\
            0 & 0 \\
            1 & 0 \\
            0 & v / L \\
        \end{array}
    \right]
    \left[
        \begin{array}{cc}
            a \\
            \tan(\delta_f) \\
        \end{array}
    \right] 
    \label{frenet system}
\end{equation}
where $ s $ and $ d $ denote the distance along the lane and the lateral offset of the lane center;
$ \mu $ is the vehicle local heading error between ego vehicle heading and closest way point orientation;
$ \kappa $ represents the curvature of the lane at the closest way point. 
On the definition of $ d $, we can use following constraints to judge if the vehicle crosses the road boundary:
\begin{equation}
    h_{r_0} = d - d_0 \geqslant 0, ~~h_{r_1} = -d - d_1 \geqslant 0          
    \label{road boundary}
\end{equation}
where $ d_0 $ and $ d_1 $ are directed distances from center of road to its boundary, normally $ d_0 + d_1 = 0 $.

Note that system \eqref{frenet system} is an affine system, and it has the same form of control input $\boldsymbol{u}$ in \eqref{veh system}. 
This means constraints of road boundary, described by limiting $d$ in \eqref{road boundary}, could be combined into the platform discussed above, though two system have different state vectors. 
 Then, the constraint of road boundary can be built following the introduced method.

\subsection{Integrating CBF}

In \ref{section percetpion conversion} and \ref{section cbf constraint}, we discuss the perception conversion method fitting with different types of senors and the CBF barrier function and constraint development method in traffic environment complying vehicle dynamics. 
Then, based on CBF theory, an optimization frame could be built:
\begin{subequations}
	\begin{align}
		\underset{\boldsymbol{u} \in \mathbb{U}}{\min} ~~ & ~~~~~~~~~~~~~(\boldsymbol{u} - \boldsymbol{u}_{o})^T Q (\boldsymbol{u} - \boldsymbol{u}_{o}) 
		\label{cost}	\\
		\text{s.t.} ~~  & ~~~~~~~~~~~~~~~~~~B(\boldsymbol{x}, \boldsymbol{u}|h_i) \geqslant 0
        \label{ob cons}\\
        & ~~ L_f h_{F_i} + L_g h_{F_i} \boldsymbol{u} + \frac{\partial h_{F_i}}{\partial \boldsymbol{x}_{ob}}\dot{\boldsymbol{x}}_{ob} + \beta_i h_{F_i} \geqslant 0
		\label{ob fs cons} \\	
             & L_{f_r}^2 h_{r_j} +  L_{f_r} L_{g_r} h_{r_j} \boldsymbol{u} + 2\gamma_j \dot{h}_{r_j} + \gamma_j^2 h_{r_j} \geqslant 0
		\label{road cons}
	\end{align}
        \label{whole cbf}
\end{subequations}
\noindent where $\boldsymbol{u}_{o}$ is original control input provided by the AD algorithm, and $Q$ is the coefficient matrix to balance the changing of acceleration and steer. Constraint \eqref{ob cons} and \eqref{ob fs cons} hopes to avoid potential collision with obstacles, where $i = 0,1,..,n-1$ and $n$ is the number of risky obstacles. Constraint \eqref{road cons} limits vehicle behavior crossing road boundaries, where $j = 0,1 $ and $f_r,g_r$ are from system \eqref{frenet system}. 

Note that the only variable the optimization frame \eqref{whole cbf} needs from the AD algorithm is original control input, since perception conversion information is enough to build constraints. This means our safe revision method has no extra requirements of types of sensors and methods used in AD algorithm, which proves the generalization of proposed method. Moreover, the optimization frame \eqref{whole cbf} still maintains quadratic planning form, so our method won't add much burden to AD system.

\section{Experiments and Validation}
\label{Section 4}

\subsection{Experiment Setting}
In this study, we implemented our proposed method on three various simulation platforms to test the effect and generalization ability on various road topologies, risk types, and different AD planning backbones. The simulators are CARLA, SUMO, and OnSite. For the CARLA simulator~\cite{dosovitskiy2017carla}, we built scenarios in RoadRunner and implemented a co-simulation platform, on which we can edit scenario logic based on various defined vehicle behavior and get feedback state from the complex vehicle dynamic system.

SUMO is a widely used simulator for continuous traffic flow~\cite{kusari2022enhancing}, implemented with a variety of default planning algorithms and map edit functions.

OnSite is a simulator that replays the traffic data collected from real-road events~\cite{zhao2022objective}. The event base is composed of interactive scenarios, each of which lasts 30 seconds approximately.

\subsection{Results}

 \subsubsection{Adaption on various planning backbones}

  To enhance the generalization ability and application value, it is crucial for the control revision method to be compatible with three various decision-making and planning backbones~\cite{wang2023differentiated, wang2024homogeneous, cheng2019longitudinal}. In this section, we test our algorithm on three simulation platforms, each of which uses an individual decision-making backbone assembled in the platform.

\begin{figure}
	\centering
	\includegraphics[width=8.5cm]{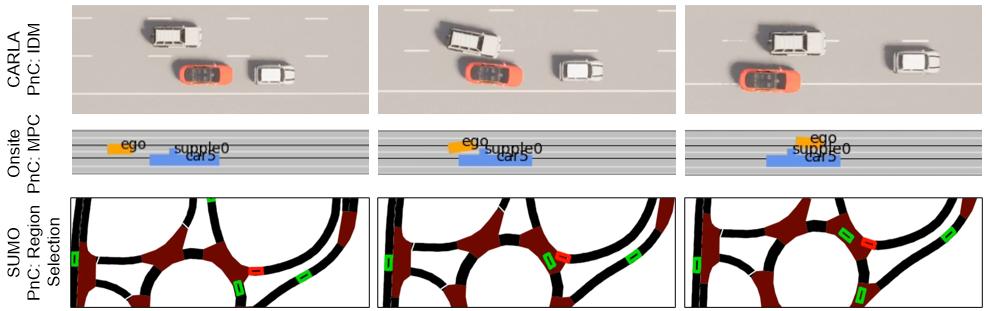}
	\caption{Adaption of the control revision module on various planning backbones and simulators.}  
	\label{Fig: various platform}
\end{figure}

  As illustrated in Fig.~\ref{Fig: various platform}, our algorithm can naturally adapt to various planning frameworks, acting as an extra safety guarantee. Even the vanilla planning method is simple and does not consider much risk assessment, by attaching our module the basic safety requirement can be satisfied.

\subsubsection{Adaption on various road topologies and risk types}

 In this part, we tested the effectiveness of the proposed control revision module on several typical risky scenarios. Comparison is also made among the AD planning with no control revision, control revision that considers obstacles only, and control revision involving obstacles and road boundary constraints. We also tested the designed control revision module on different road structures, such as straight roads, curve roads, ramps, etc.
 
\begin{figure}
	\centering
	\includegraphics[width=8.5cm]{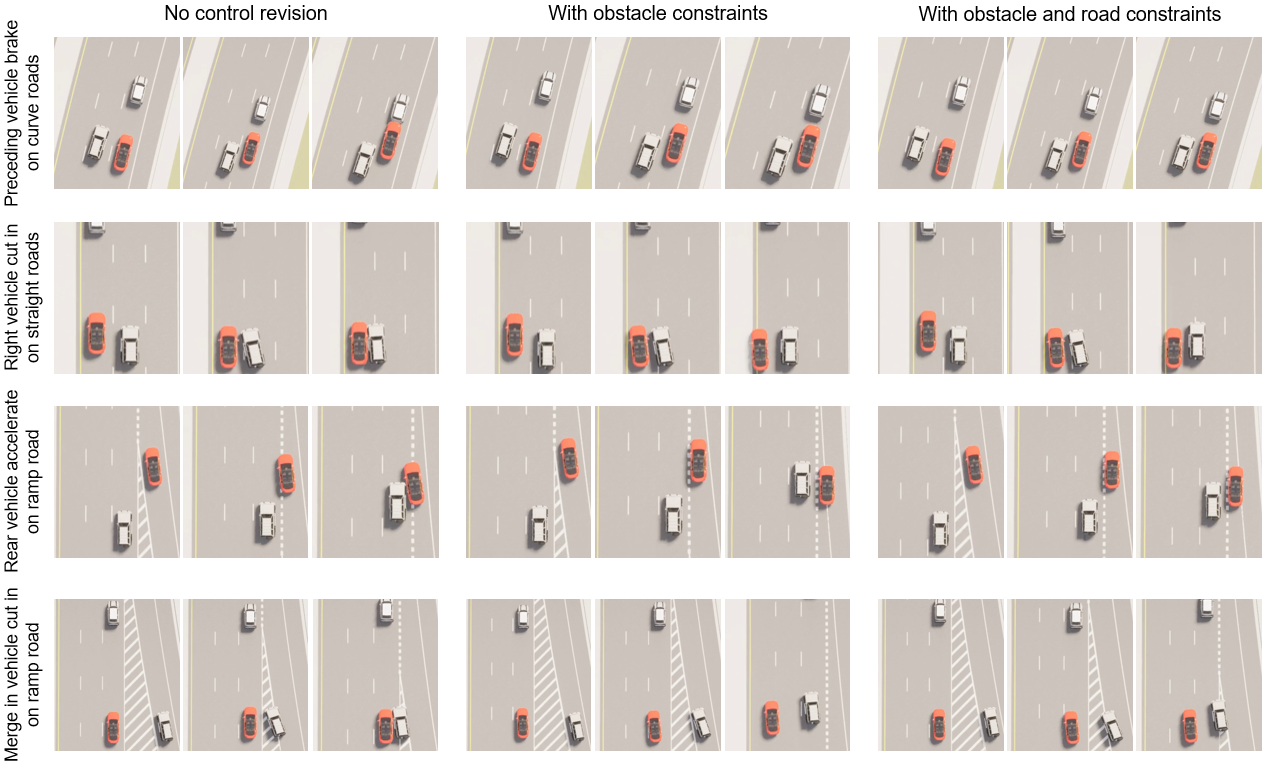}
	\caption{Experiment results on various kinds of topologies and scenarios}  
	\label{Fig: Carla results}
\end{figure}

\begin{figure}
	\centering
	\includegraphics[width=8.5cm]{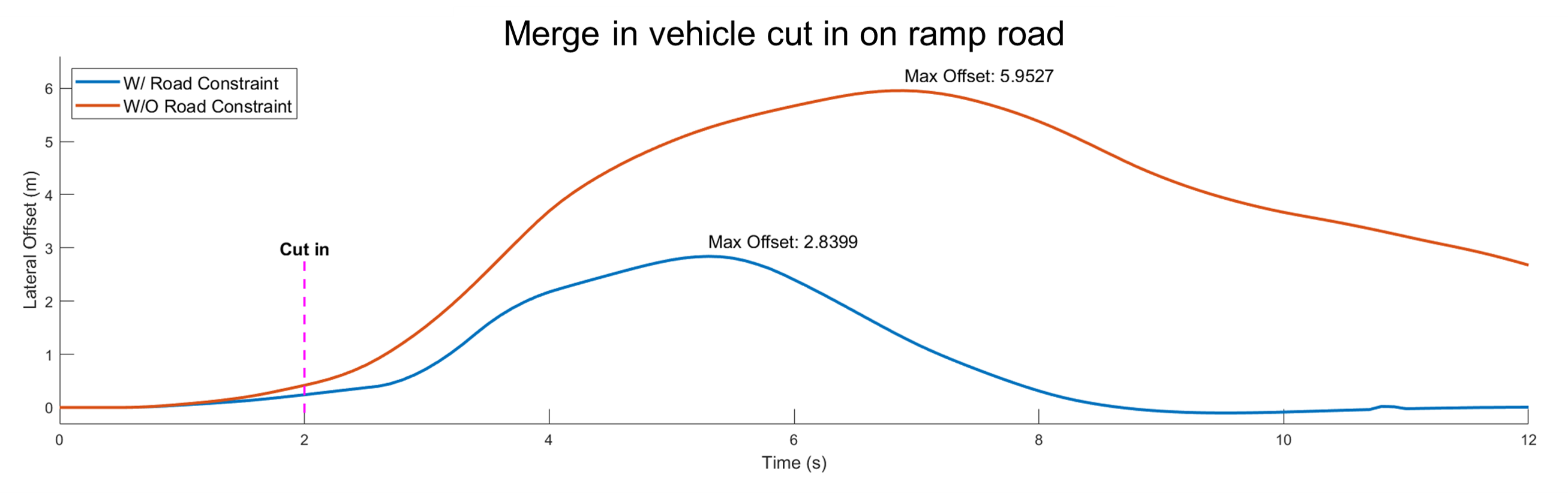}
	\caption{Comparison of lateral offset during safe control revision.}  
	\label{Fig: lateral offest}
\end{figure}

 As demonstrated in Fig.~\ref{Fig: Carla results}, the red vehicle is the controlled agent equipped with autonomous planning algorithms, and the white ones are disturbance vehicles. In the first scene when the preceding vehicle suddenly brakes, the ego vehicle could respond timely after applying control revision. In the second case, the right vehicle suddenly cut in. When no road constraints were considered, although the ego vehicle could avoid collision, it directly drove on the road curb and generated new risks. Comparatively, by implementing our method, the ego vehicle escaped from the risk by minor lateral displacement and remained in road regions. Similarly, in the two ramp scenarios, facing the threats from behind and left, our proposed method can avoid the crash in a more stable manner. Fig.~\ref{Fig: lateral offest} demonstrates the lateral offset of the control revision with and without in the fourth scene. By considering lane and road boundary restrictions, the maximum lateral offset decreased from 5.95m to 2.84m.

 In addition, since the road boundary constraints are established based on the Frenet Coordinate system, we can cope with various road topologies by one unified method.

 % \subsubsection{Test on various danger types}

 % As demonstrated in Fig.X, in the ramp merge in TODO: scene and performance description.
 
 % As for the comparison, when no revision is applied, severe crashes will happen causes by the sudden danger behavior from the disturbance vehicle. After considering obstacle constraints, other the vehicle-vehicle accident can be avioded, the ego vehicle tends to largely deviate from the original lane, generating new danger because of driving out of road boundary. When considering both the obstacle and road boundary constraints, the ego vehicle could avoid the collision by a small lateral displacement so that it can stay within the drivable area.

 % Through the unified modeling of various traffic elements, the control safety revision module can ensure collision avoidance and cope with different driving danger genres.

 \subsubsection{Test on continuous traffic flow scenes}
 
\begin{figure}
	\centering
	\includegraphics[width=5.5cm]{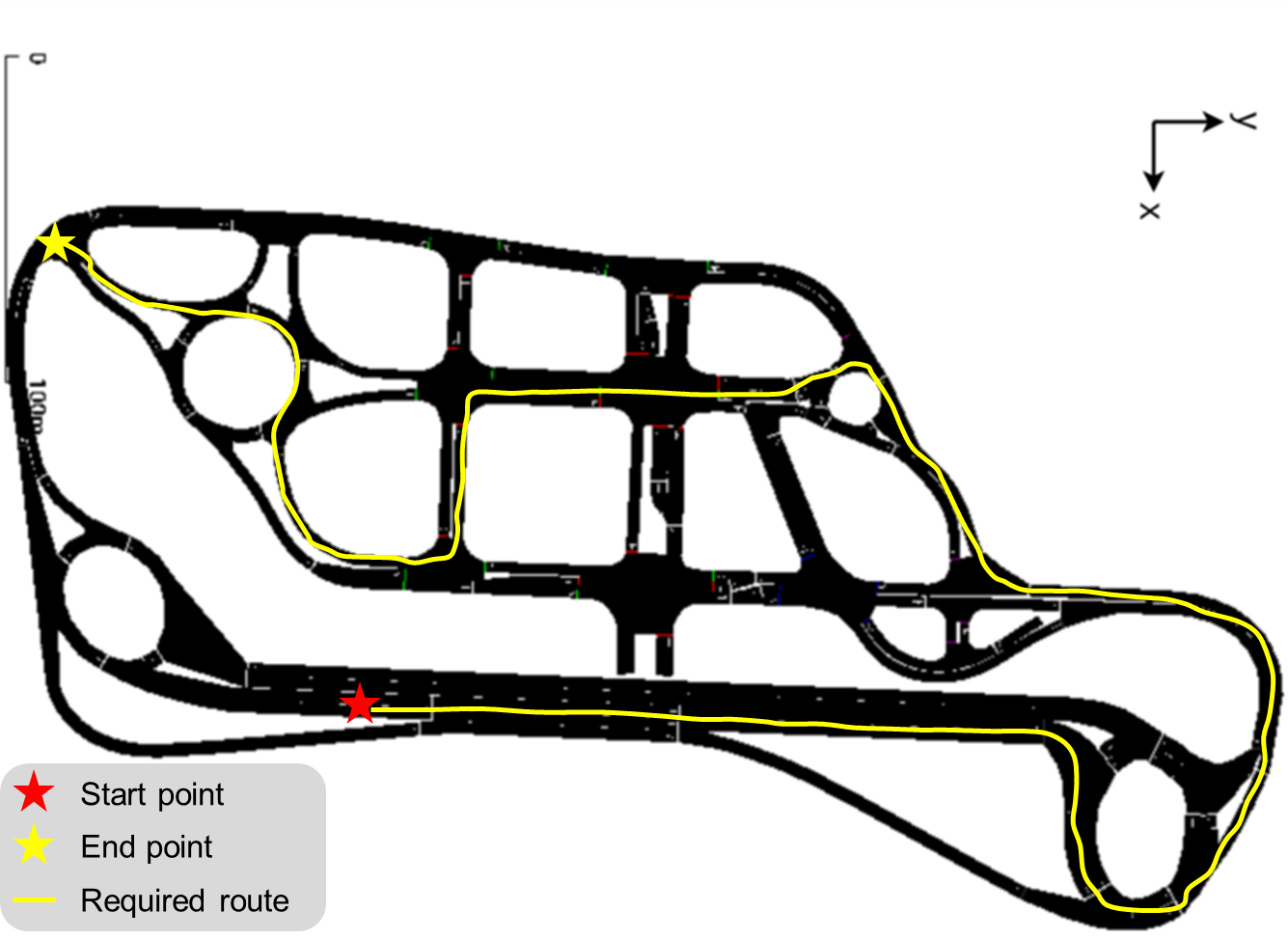}
	\caption{The scene map of continuous traffic flow experiment.}  
	\label{Fig: sumo map}
\end{figure}

 In this part, we record the accident rate of the AD planning system with and without control control revision on the SUMO platform. The testing route is approximate 1.1 km long including various types of traffic scenes, as shown in Fig.~\ref{Fig: sumo map}. The task of the ego vehicle is to follow the yellow route until reaches the final point. Apart from the controlled ego vehicle, background vehicles are randomly and continuously generated on all lanes, which are designed to act aggressively for posing severe challenges on driving safety.  For the baseline and our method, we conduct 10 times of experiments.

 The results are shown in Table.~\ref{Tab:sumo result}. Under the aggressive traffic background, when no control revision is equipped all of the cases encountered collision accidents, and after implementing the proposed module the accident number was reduced to 4. To further investigate safety performance, we refer to the definition of~\cite{wu2021mid}, dividing the collisions into at-fault accidents where the ego vehicle should be responsible for the crash and the passive ones where the ego vehicle has no fault. According to the results, the original at-fault accident number is 3, and after implementing the revision module all these collisions are avoided.  

 \begin{table}
    \centering
    \caption{Comparison results on continous traffic flow scenes}
    \begin{tabular}{ccc}
        \toprule Method & Accident number & At-fault accident number \\
        \midrule
        w/o control revision & 10/10 & 3/10 \\
        w control revision & 4/10 & 0/10 \\
        \bottomrule
    \end{tabular}
    \label{Tab:sumo result}
\end{table} 

\subsubsection{Ablation study of perception data conversion}

\begin{figure}
	\centering
	\includegraphics[width=8.5cm]{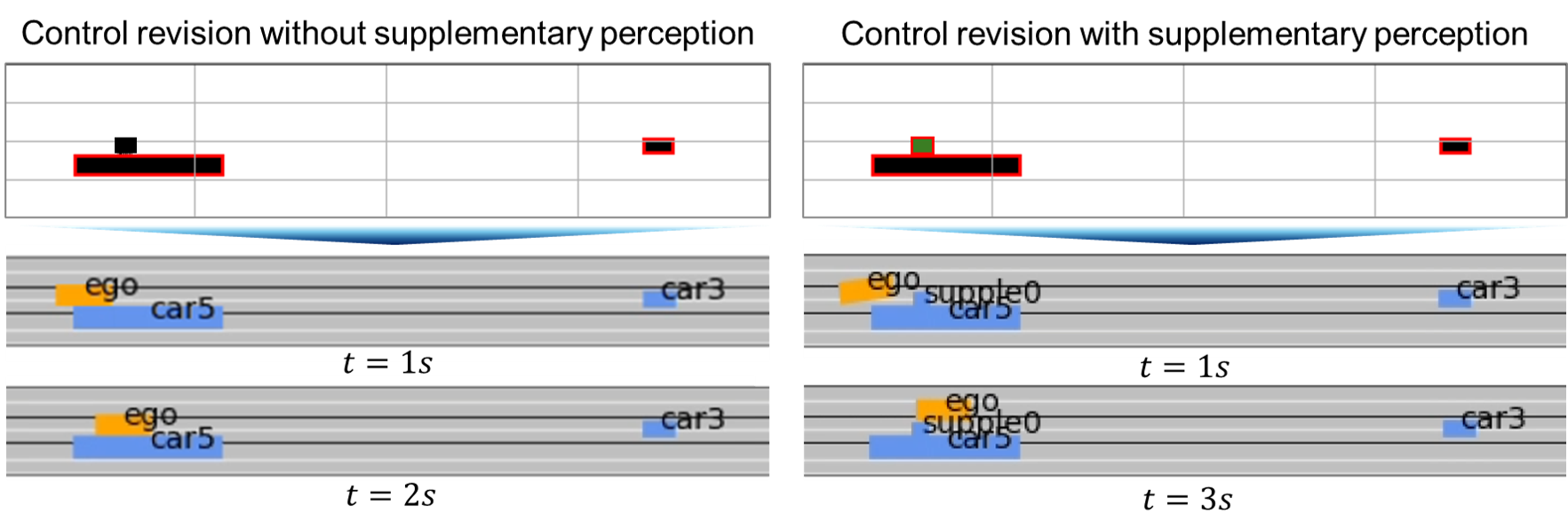}
	\caption{Ablation results on the perception data conversion.}  
	\label{Fig: supple percp result}
\end{figure}

To further illustrate the effect of the proposed perception data conversion, we compare the control revision module with and without the supplementary bounding boxes from occupancy grid map on the OnSite simulator platform. As demonstrated in Fig.~\ref{Fig: supple percp result}, the black rectangles with red frame lines are the vanilla vectorized perception inputs, and the green rectangle with red frame lines are the supplementary bounding box derived from the process introduced in \ref{section percetpion conversion}, which is an object extending out of the truck body. After applying the supplementary perception, the revision module could avoid collision with the irregular obstacle which wasn't covered by vanilla perception. By adopting both vectorized bounding box and occupancy grid map, our proposed revision method becomes more complete and comprehensive, further guaranteeing driving control safety.

\subsection{Real-world Platform Validation}

\begin{figure}
	\centering
	\includegraphics[width=8.5cm]{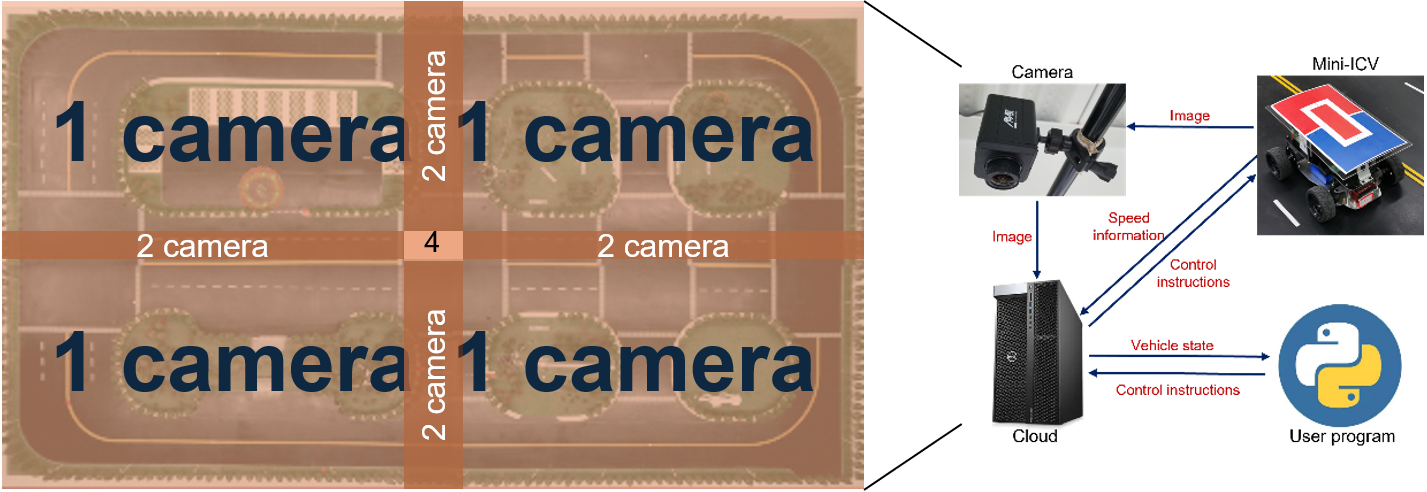}
	\caption{An introduction of MCCT platform.}  
	\label{Fig: Scity details}
\end{figure}

To further demonstrate the feasibility of the proposed control revision method, we validate the algorithm on a scale physical platform named MCCT, the constructions of which are shown in Fig.~\ref{Fig: Scity details}~\cite{dong2023mixed}. Four Bird-eye view cameras recognize the mini-vehicles' physical properties on a scaled traffic sandbox. The cloud server acts as an information hub that exchanges the perception data from cameras and control commands from users' programs, enabling the control revision module to be implemented on real-vehicle platforms.

\begin{figure}
	\centering
	\includegraphics[width=8.5cm]{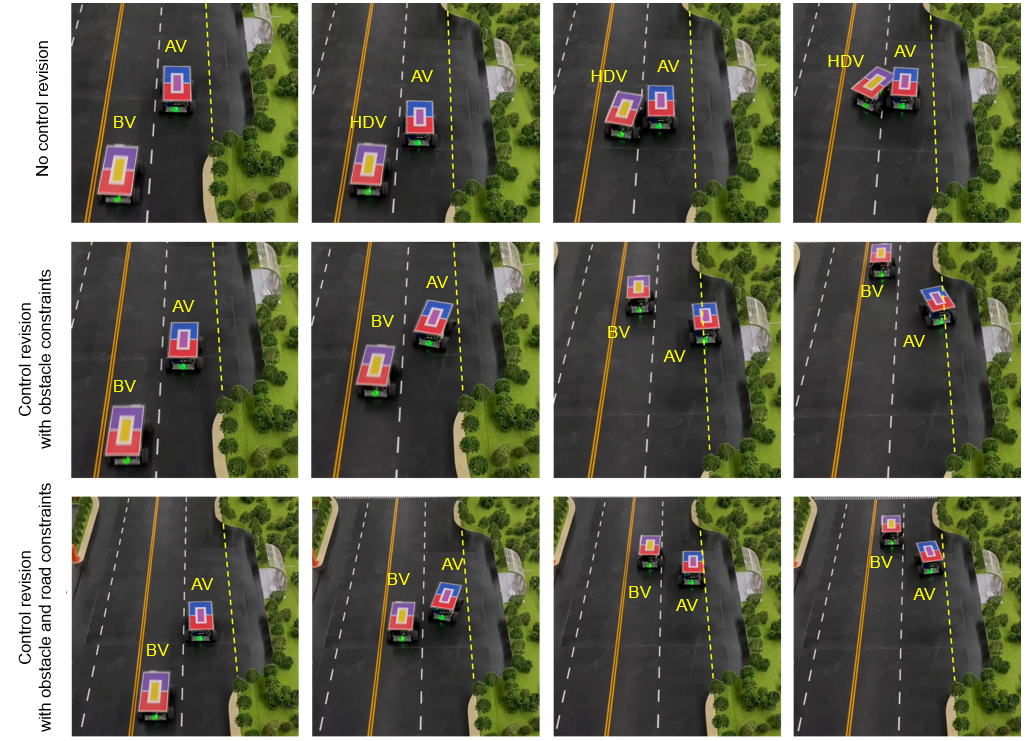}
	\caption{Validation results on MCCT platform.}  
	\label{Fig: Scity validation}
\end{figure}

We compare the control results of three modes: with no control revision, control revision with only obstacle constraints, and control revision with both obstacle and road boundary constraints, as illustrated in Fig.~\ref{Fig: Scity validation}. In this experiment, the left vehicle was set as a background vehicle (BV) which generated dangerous behaviors, and the right one was the autonomous driving vehicle (AV) controlled by the server. The yellow dashed line on the right side of the AV was considered as the virtual road boundary. When no control revision was implemented, the AV did not react to the BV's cutting in, leading to a collision. Comparatively, the AV could avoid the crash with BV by driving to the right side after using a naive CBF revision. However, since the farther away from BV the safer it would be, the AV manipulated a large lateral displacement to escape without considering road boundary restrictions. Hence, a complete traffic element constraint representation involving both obstacles and road boundaries is indispensable for control revision. After considering both factors, when BV cut in, the AV managed to avoid the collision by conducting a minor lateral escape while remaining the main body within a drivable region, reaching a balance between obstacle risk and road boundary risk.

The validation on the physical platform verifies the effectiveness and real-time feasibility of the proposed control revision.

\section{CONCLUSIONS}
\label{Section 5}

This paper proposes a generalized control revision method for autonomous driving safety. Based on a perception data conversion layer and unified traffic element constraint representation, the control revision module can ensure autonomous vehicle control safety comprehensively. The proposed method was tested on multiple simulators including CARLA, SUMO, and OnSite and validated on a real-world platform MCCT. Experiments proved that our method could revise the danger behaviors and avoid accidents with various kinds of perception data and planning backbone under different road topologies. Tests on continuous traffic flow scenes also showed that the accident rate can be greatly reduced by applying control revision. Validations on real-world platforms verified the feasibility.

\addtolength{\textheight}{0cm}   % This command serves to balance the column lengths

%%%%%%%%%%%%%%%%%%%%%%%%%%%%%%%%%%%%%%%%%%%%%%%%%%%%%%%%%%%%%%%%%%%%%%%%%%%%%%%%

%%%%%%%%%%%%%%%%%%%%%%%%%%%%%%%%%%%%%%%%%%%%%%%%%%%%%%%%%%%%%%%%%%%%%%%%%%%%%%%%



% \section*{ACKNOWLEDGMENT}

% The preferred spelling of the word ÒacknowledgmentÓ in America is without an ÒeÓ after the ÒgÓ. Avoid the stilted expression, ÒOne of us (R. B. G.) thanks . . .Ó  Instead, try ÒR. B. G. thanksÓ. Put sponsor acknowledgments in the unnumbered footnote on the first page.

\bibliography{IEEEabrv, mybibfile}

\end{document}